\documentclass[runningheads]{llncs}

\usepackage{eccv}

\usepackage{eccvabbrv}

\usepackage{graphicx}
\usepackage{booktabs}

\usepackage[accsupp]{axessibility}  % Improves PDF readability for those with disabilities.

\usepackage{hyperref}

\usepackage{orcidlink}

\definecolor{cvprblue}{rgb}{0.21,0.49,0.74}
\usepackage{diagbox}
\usepackage{colortbl} 
\definecolor{mygray}{gray}{.9}
\makeatletter
\newcommand{\thickhline}{%
    \noalign {\ifnum 0=`}\fi \hrule height 1pt
    \futurelet \reserved@a \@xhline
}

\usepackage{wrapfig}

\begin{document}

% ---------------------------------------------------------------
% TODO REVIEW: Replace with your title
\title{Radiance Field Learners As UAV \\ First-Person Viewers} 

% TODO REVIEW: If the paper title is too long for the running head, you can set
% an abbreviated paper title here. If not, comment out.
\titlerunning{Radiance Field Learners As UAV First-Person Viewers}

% TODO FINAL: Replace with your author list. 
% Include the authors' OCRID for the camera-ready version, if at all possible.
\author{Liqi Yan\inst{1}\orcidlink{0000-0002-7077-4947} \and
Qifan Wang\inst{2} \and
Junhan Zhao\inst{3} \and
Qiang Guan \inst{4} \and
Zheng Tang \inst{5} \and
Jianhui Zhang \inst{1} \and
Dongfang Liu* \inst{6}
}
% TODO FINAL: Replace with an abbreviated list of authors.
\authorrunning{Liqi Yan, et al.}
% First names are abbreviated in the running head.
% If there are more than two authors, 'et al.' is used.

% TODO FINAL: Replace with your institution list.
\institute{Hangzhou Dianzi University, Hangzhou, Zhejiang 310024, China \and Meta AI, Menlo Park, CA 94025, USA \and Harvard University, Boston, MA 02138, USA \and Kent State University, Kent, OH 44240, USA \and  NVIDIA, Redmond, WA 98052, USA \and Rochester Institute of Technology, Rochester, NY 14623, USA \\
\email{* is the contact author }
}

\maketitle

% \begin{abstract}
%   The abstract should concisely summarize the contents of the paper. 
%   While there is no fixed length restriction for the abstract, it is recommended to limit your abstract to approximately 150 words.
%   Please include keywords as in the example below. 
%   This is required for papers in LNCS proceedings.
%   \keywords{First keyword \and Second keyword \and Third keyword}
% \end{abstract}

\begin{abstract}
First-Person-View (FPV) holds immense potential for revolutionizing the trajectory of Unmanned Aerial Vehicles (UAVs), offering an exhilarating avenue for navigating complex building structures. Yet, traditional Neural Radiance Field (NeRF) methods face challenges such as sampling single points per iteration and requiring an extensive array of views for supervision. UAV videos exacerbate these issues with limited viewpoints and significant spatial scale variations, resulting in inadequate detail rendering across diverse scales. In response, we introduce \textbf{FPV-NeRF}, addressing these challenges through three key facets: (1) \textbf{Temporal consistency}. Leveraging spatio-temporal continuity ensures seamless coherence between frames; (2) \textbf{Global structure}. Incorporating various global features during point sampling preserves space integrity; (3) \textbf{Local granularity}. Employing a comprehensive framework and multi-resolution supervision for multi-scale scene feature representation tackles the intricacies of UAV video spatial scales. Additionally, due to the scarcity of publicly available FPV videos, we introduce an innovative view synthesis method using NeRF to generate FPV perspectives from UAV footage, enhancing spatial perception for drones. Our novel dataset spans diverse trajectories, from outdoor to indoor environments, in the UAV domain, differing significantly from traditional NeRF scenarios. Through extensive experiments encompassing both interior and exterior building structures, FPV-NeRF demonstrates a superior understanding of the UAV flying space, outperforming state-of-the-art methods in our curated UAV dataset. Explore our project page for further insights: \url{https://fpv-nerf.github.io/}.
\keywords{Computer Vision \and Spatial Perception \and Neural Radiance Field \and First-Person View (FPV) \and Unmanned Aerial Vehicle (UAV)}
\end{abstract}    
\section{Introduction}
\label{sec:intro}

\begin{figure}[t]
  \centering
   \includegraphics[width=1.0\linewidth]{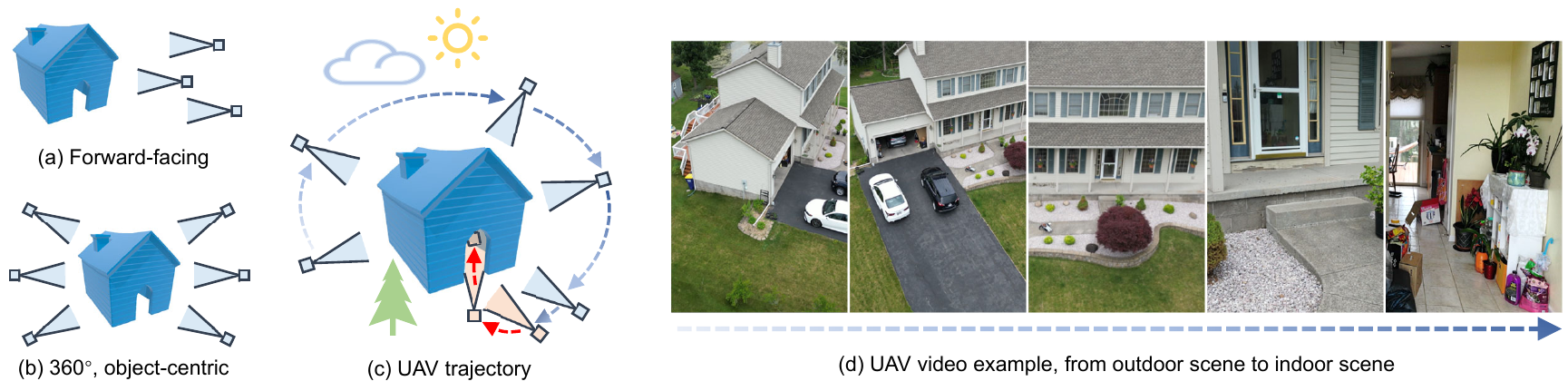}
   \vspace{-3mm}
   \caption{\textbf{Comparison of our proposed FPV-NeRF and previous NeRF-based methods.} Previous NeRF can be divided into two types: forward-facing and 360° object centric. 
   % In UAV videos, the trajectory contains both of them and should consider different scales (including both outside and inside of the building). 
   In UAV videos, view synthesizing faces the following challenges: 1) Degree of view restriction, as UAV perspectives are limited by drone trajectories; and 2) Scene change, as UAVs encounter significant changes in scene scale and lighting conditions when transitioning from outdoors to indoors. }
   \label{fig:intro}
   \vspace{-0.5cm}
\end{figure}

In computer vision, advancements in Robotic Navigation, including Visual-SLAM \cite{mur2017orb} and MVS \cite{goesele2006multi}, alongside innovations in Computer Graphics, such as Novel View Synthesis \cite{mildenhall2019local}, have become integral to applications like environmental monitoring and disaster response \cite{cui2021tf,liu2021densernet,yan2023prompt,lu2023transflow}. These needs establish Neural Radiance Field (NeRF) \cite{mildenhall2021nerf} as an emerging paradigm, using Multi-layer Perceptron networks for scene representation and excelling in rendering high-quality images from novel camera poses.

Despite NeRF’s success \cite{sun2022direct,muller2022instant,barron2022mip,turki2022mega,tancik2022block}, \textbf{it encounters challenges in UAV-captured multi-scale videos} due to its reliance on a single ray from a pixel in training, limiting its effectiveness. Firstly, the model lacks temporal consistency, resulting in non-smooth sequences during novel view generation. Secondly, NeRF struggles to construct a comprehensive global structure, particularly in low-texture regions like solid-colored surfaces or grassy fields. Thirdly, difficulties arise in rendering local details when original video perspectives are insufficient, especially around structures like buildings. Consequently, when applied to generate novel First-Person-View perspectives from complex UAV trajectories, such as zooming from exterior to interior (Fig.~\ref{fig:intro}), NeRF's quality diminishes, highlighting its limitations in handling diverse scenarios and viewpoints.

Building on the preceding discussions, we introduce \textbf{FPV-NeRF}, a \underline{F}irst-\underline{P}erson \underline{V}iew synthesis framework tailored for UAVs, leveraging the power of Neural Radiance Field. Drones equipped with a first-person view (FPV) provide immersive videos, offering a unique perspective that places viewers directly in the midst of the action. Extensive research has highlighted that humans grasp a comprehensive spatial understanding from FPV videos, mentally transitioning between first and third perspectives with a holistic grasp of the environment. If a UAV can transform a third-person view into first-person, it demonstrates spatial cognition akin to humans. Our FPV-NeRF serves as a robust and interpretable backbone for UAVs, excelling in multi-scale spatial structure recognition \cite{zhang2023global, gao2019res2net} while acknowledging and surpassing the limitations of existing NeRF.

Specifically, FPV-NeRF divides the entire airspace into regions, recursively breaking them into subregions. This multi-scale space constructor dynamically adjusts the three-dimensional model composition based on the drone's flight trajectory, considering both global and local perspectives. Unlike previous NeRFs relying solely on MLP layers for color prediction, FPV-NeRF optimizes global-local features for all points in the estimated UAV flying space. Our proposed method includes: \textbf{I)} Multi-scale camera space estimation, focusing on trajectory reconstruction using adjacent time frames to model environmental space. The trajectory space undergoes division, applying distinct coordinate distortion functions based on the UAV's position and pose in adjacent frames. \textbf{II)} First-person view video generation with a global-local scene encoder: \textbf{i)} a learnable volume embedding indicating general features of each block; \textbf{ii)} a point location embedding providing continuous features based on position, enhancing multi-scale encoding with various frequency components; \textbf{iii)} a level embedding offering global features for resolution rendering and cross-attention between levels. \textbf{III)} For training, we propose a comprehensive loss function with three terms, incorporating Optimal Transport for disparity alignment and bounded variation for smoothness consistency among neighboring regions. 

Essentially, FPV-NeRF embodies several appealing qualities: \textbf{First}, it achieves improved \textbf{\textit{temporal consistency}}. The robust correlation timing mechanism and spatial coherence between adjacent frames in video sequences empower FPV-NeRF to construct a more continuous and smooth overall structure. \textbf{Second}, it enhances the integrity of the \textbf{\textit{global structure}}. Through the thoughtful design of cross-attention between different resolutions and the acquisition of a volume feature, the training of each point's feature considers the context of surrounding points, mitigating the risk of overlooking low contextual elements. \textbf{Third}, it excels in providing \textbf{\textit{local granularity}}. The subdivision of space into regions with distinct warping functions, coupled with the transformation of coordinate positions of sampling points into high-dimensional features, including high-frequency components, ensures the high-fidelity restoration of local details.
FPV-NeRF is an intuitive yet general classification framework; it is compatible with different NeRF-based first-person view generation methods. We experimentally show: \textbf{In \S \ref{sec:comparison_sota}}, FPV-NeRF outperforms parametric counterparts, \textit{i.e.}, \textbf{1.61-7.77} PSNR and \textbf{0.021-0.145} SSIM, on our proposed UAV dataset. With voxel-based architecture, FPV-NeRF also outperforms those SOTA methods on previous frequently used NeRF datasets, \textit{i.e.}, \textbf{0.03-3.07} PSNR on NeRF-360-V2 dataset and \textbf{0.01-1.65} PSNR on LLFF dataset. \textbf{In \S \ref{sec:ablation_study}}, our multi-scale camera space estimation method enhances UAV dataset performance by \textbf{1.52-5.23} PSNR. Furthermore, our global-local encoder significantly boosts performance (\textbf{1.12-4.36} PSNR), and the incorporation of our comprehensive loss further elevates performance (\textbf{1.49-4.85} PSNR).

\vspace{-0.2cm}
\section{Related Works} 
\textbf{Spatial Perception for UAVs}
Spatial perception, crucial in applications like indoor mapping, drones, and self-driving vehicles, has been extensively studied \cite{schueftan2015indoor, von2017monocular, qin2020avp}. Technological progress, particularly in cameras and LiDAR, has accelerated the adoption of SLAM systems. Visual SLAM \cite{macario2022comprehensive, sumikura2019openvslam, teed2021droid, wang2020tartanair} is widely employed for its compatibility with low-cost vision sensors. These systems, utilizing graph optimization, can be categorized as sparse feature-based or direct methods. Sparse feature-based methods track feature points to generate a 3D map \cite{klein2007parallel, mur2017orb}, while direct methods minimize pixel intensity differences for 3D camera movement and environment estimation \cite{fontan2020information, engel2014lsd}. Despite popularity, Visual SLAM faces accuracy limitations, hindering precise navigation \cite{yan2021hierarchical,yan2020multimodal}. Current research increasingly incorporates novel view generation to achieve spatial perception. Unlike traditional methods, novel view generation accurately reconstructs UAV navigation, bringing it closer to the real environment. \\
\textbf{First-Person View Generation Methods}
Novel first-person view synthesis involves two primary methods. Generative models, as seen in recent works like \cite{tseng2023consistent, wynn2023diffusionerf, chan2023generative}, use the diffusion model to address the challenge of generating random yet plausible samples from the conditional distribution. However, these methods, relying on generative priors, may lack an inherent understanding of the spatial environment \cite{chan2023generative}. Alternatively, some methods utilize 3D reconstruction \cite{mescheder2019occupancy} by estimating geometry from images, employing representations like point clouds, depth maps, meshes, or volumetric implicit functions \cite{furukawa2015multi, park2019deepsdf}. Traditional multi-view stereo (MVS) techniques and surface reconstruction methods (\textit{e.g.}, Poisson \cite{kazhdan2006poisson, kazhdan2013screened}, Delaunay triangulation \cite{labatut2009robust}) have evolved, with recent learning-based depth estimation methods exhibiting exceptional performance \cite{darmon2021deep, gu2020cascade, huang2018deepmvs, liu2019neural, yao2018mvsnet, yao2019recurrent}. Despite their success in generating 3D models, these methods depend heavily on accurate depth estimation: diffusion models are typically employed for generating 2D images and often necessitate accurate depth estimation or NeRF to extend its application to 3D space \cite{tseng2023consistent, wynn2023diffusionerf, chan2023generative}, and MVS methods rely on GT depth maps to compute the loss of the prediction \cite{darmon2021deep, huang2018deepmvs}. Consequently, the rise of Neural Radiance Field (NeRF) \cite{mildenhall2021nerf} has positioned it favorably for novel view synthesis, encompassing depth information in an implicit embedding with heightened accuracy through iterative training processes. \\
\textbf{Neural Radiance Field (NeRF)}
In contrast to traditional 3D reconstruction techniques, NeRF \cite{mildenhall2021nerf} employs implicit neural network features for spatial representation, efficiently generating novel view images. Since its introduction, NeRF has spurred research in neural representations for tasks like novel view synthesis \cite{sitzmann2019scene, tancik2023nerfstudio}, relighting \cite{boss2021nerd, munkberg2022extracting, zhang2021physg, zhang2021nerfactor}, generalization to new scenes \cite{chen2021mvsnerf, liu2022neural, suhail2022generalizable, wang2021ibrnet}, shape representation \cite{muller2022instant, takikawa2021neural}, and multiview reconstruction \cite{oechsle2021unisurf, wang2021neus, yariv2021volume, yariv2020multiview}. These methods implicitly reconstruct scenes using multi-layer perceptrons (MLPs), yielding impressive novel view synthesis outcomes, even with very few training views \cite{yu2021pixelnerf, yang2023freenerf}.  Furthermore, recent methodologies exhibit promise in tackling the intricacies associated with large-scale or unbounded scenes \cite{barron2022mip,turki2022mega,tancik2022block,xiangli2022bungeenerf}. However, existing NeRF approaches like Instant-NGP \cite{muller2022instant}, TensoRF \cite{chen2022tensorf}, DVGO \cite{sun2022direct} and Mip-NeRF 360 \cite{barron2022mip} are designed primarily for forward-facing or 360° object-centric trajectories for a fixed scale, displaying limited adaptability to multi-scale UAV videos. Besides, prevailing methodologies exclusively rely on isolated sampled point features per iteration, neglecting global features. This confines them to singular trajectories and scales, limiting UAV video exploration. In contrast, our proposed method is tailored for tasks involving both forward-facing and 360° object-centric trajectories, aiming to ensure consistent, high-quality long-term view synthesis for UAVs.

\begin{figure}[t]
  \centering
   \includegraphics[width=1.0\linewidth]{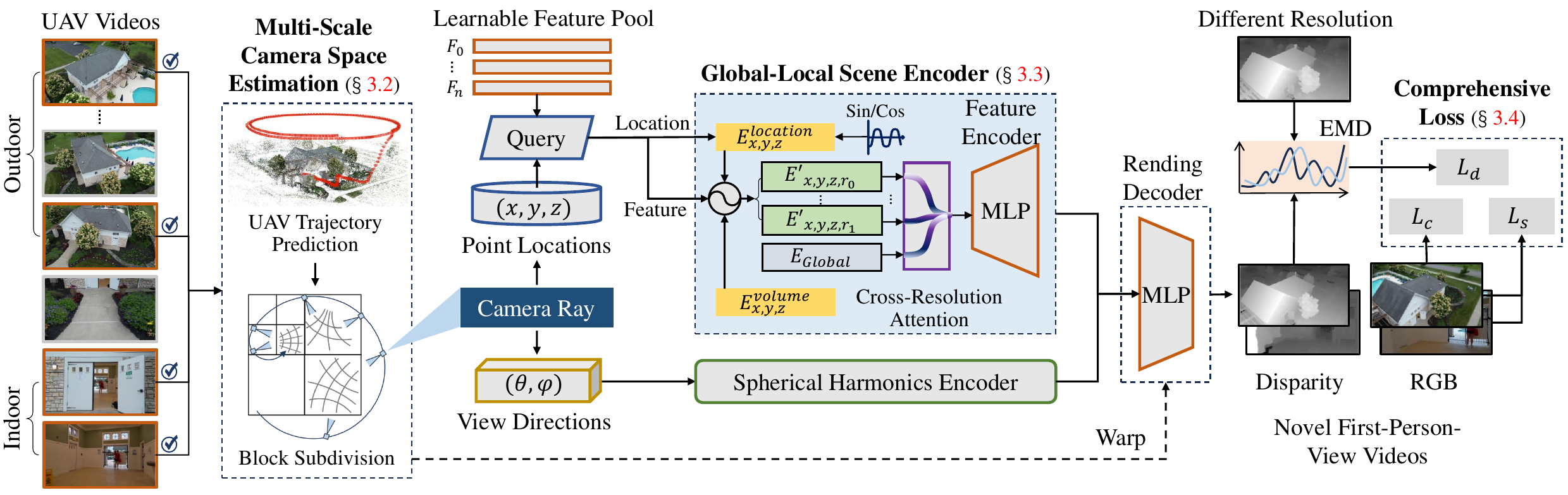}
   \caption{\textbf{The overall framework of our method.} After estimating the camera location and pose space using various wrapping functions, we can sample a pixel in a frame as a view ray, which consists of a sequence of point positions in this estimated space. During training, we use these point positions to query their learnable features from a feature pool. Then, we pass those point positions and corresponding features through a global-local encoder and rendering decoder to obtain the predicted color for this pixel. Comparing this predicted pixel color with the ground truth color of this pixel can supervise the network. During testing, we input a novel auto-generated point position sequence into this pipeline and can finally obtain a novel first-person view video (see Fig.~\ref{fig:vis_trajectory}).} 
   \label{fig:framework}
   \vspace{-0.3cm}
\end{figure}

\section{Methods}
\vspace{-0.1cm}
Given a video $V$ with a length of $L$, without any location and pose information captured from UAV, the goal of FPV-NeRF is to reconstruct a radiance field of the scene to generate a novel first-person view video. In the following, we first give an overview of FPV-NeRF framework.

\vspace{-0.2cm}
\subsection{Overview} \label{sec:framework}
The FPV-NeRF framework is delineated as follows:
I) \textbf{Multi-scale camera space estimation (\S \ref{sec:space_estimation})}: In the initial step, illustrated in Fig.~\ref{fig:framework}, keyframes are selected to predict the trajectory and poses of the UAV camera. These predictions occur in a subdivided space with diverse warping functions, using a Jacobian matrix for seamless point warping between 3D and multi-camera 2D spaces. A learnable feature pool, functioning as a neural embedding, captures inherent features of each point in this spatial configuration.
II) \textbf{Global-local scene encoder (\S \ref{sec:encoder})}: Following the approach of \cite{muller2022instant}, we synthesize images at various resolutions within the spatial domain, exploring global-local information across resolutions. For each pixel in the synthesized image, camera rays are traced through the scene, generating sampling points. The global-local scene encoder utilizes point location information and queried features at these sampling points to compute hidden features.
III) \textbf{Rendering and comprehensive loss (\S \ref{sec:loss})}: Following the scene encoder, a rendering MLP layer predicts local color and density for each ray, considering point features and camera viewing direction. Volume rendering techniques generate the image from computed colors and densities. The loss is computed on the rendered image, disparity, and ground truth image, providing a holistic evaluation of the model.

\begin{figure}[t]
  \centering
   \includegraphics[width=1.0\linewidth]{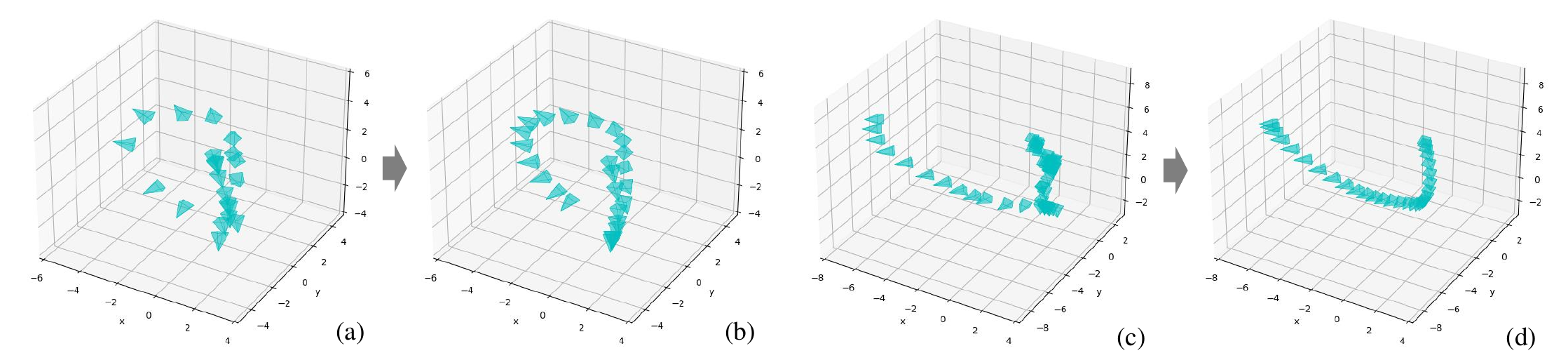}
   \caption{\textbf{Visualization of UAV trajectory and auto-generated FPV video trajectories with camera poses.} (a-b) \textit{Tentage} scene. (c-d) \textit{Market} scene.}
   \label{fig:vis_trajectory}
   \vspace{-0.5cm}
\end{figure}

%-------------------------------------------------------------------------
\subsection{Multi-Scale Camera Space Estimation} \label{sec:space_estimation}
We initially establish a UAV trajectory space to define camera positions and orientations within its spatial disposition:

\textbf{Keyframe selector.} We compute information entropy \cite{shannon1948mathematical} for each UAV video frame, considering color entropy within frames and entropy between frames. These components are multiplied to assess frame information, expressed as:
\begin{align}
    & I^{(l)} = -\sum_{i=0}^{255} p_i^{(l)} log p_i^{(l)} \times \frac{1}{L} \sum_{k=0}^{L}\sum_{i=0}^{255} \sigma \left(p_i^{(k)}, p_i^{(l)} \right), \\
    & \sigma \left(p_i^{(k)}, p_i^{(l)} \right) = \left|p_i^{(k)} log p_i^{(k)} - p_i^{(l)} log p_i^{(l)} \right|,
\end{align}
where $I^{(l)}$ is the information entropy of the $l$-th frame (among $L$ frames), and $p_i$ is the probability of the $i$-th grayscale value from the histogram. Keyframes with notable information entropy in both aspects are identified.

\textbf{UAV trajectory prediction.}  In contrast to previous Structure from Motion (SfM) methods \cite{schonberger2016structure}, our method capitalizes on sequential video acquisition, where frames naturally exhibit visual overlap, eliminating the need for exhaustive image pair matching. Employing loop detection based on a vocabulary tree \cite{schonberger2017vote}, each image is matched with its temporally closest counterpart.  Suppose a scene part contains $M$ points $S=\{\delta_0,...\delta_M\}$, for each point $\delta_j \in S$ appearing for a duration of $T_i$ in the video, considering the $l_j$-th frame is where this point initially appears, the sequential trajectory can be predicted by minimizing the reprojection error:
\begin{equation}
    \arg \min_{S, \Theta} \sum_{j=0}^{M} \sum_{t=l_j}^{l_j+T_j} \bigg|\bigg|q_{t,j} - P\left( \Theta_{t}, \delta_{j} \right)\bigg|\bigg|,
\end{equation}
where $\Theta_{t}$ is the camera's internal and external parameters at time $t$, and $q_{t,j}$ is the pixel location of $\delta_j$ in that frame. $P$ is the project function.
Output forms Euclidean space for camera matrices, processed by a subsequent block subdivision module for multi-block warping camera space.

\textbf{Block subdivision for distinct warping.} The core requirement for spatial warping involves transforming the original Euclidean space, necessitating axis-aligned grid establishment within the warped space for alignment with camera rays. Diverging from prior NeRF approaches \cite{mildenhall2021nerf,barron2022mip} reliant on a single warping function, our scenes exhibit varied camera trajectories across regions (depicted in Fig.~\ref{fig:framework}). This mandates crafting distinct warping functions (i.e., Jacobian matrices) for each region, ensuring precise alignment of camera position and direction with region-specific pixels. Utilizing an octree structure for region subdivision, our goal is to identify visible cameras intersecting with a tree node of side length $s$. If a visible camera center is within a distance $d \leq \lambda s$ (with $\lambda$ set at 3), the node undergoes subdivision into 8 child nodes; otherwise, it becomes a leaf node. This process repeats until all leaf nodes are obtained.

\subsection{Global-Local Scene Encoder} \label{sec:encoder}
After obtaining the point embedding $E_{x,y,z}$ by querying the point within the feature pool and applying Multiresolution Hash Encoding \cite{muller2022instant} to obtain the preprocessed point embedding $E_{x,y,z}'$, we then input $E_{x,y,z}'$ into the Global-Local Scene Encoder to compute a holistic feature representation for each point.
% This encoder encompasses several novel components, including a newly introduced cross-resolution attention mechanism, a novel volume feature integration, and a unique position embedding.

\textbf{Cross-resolution attention.} Scene features at various scales are effectively captured through multi-resolution levels. Global aspects are accentuated at low resolution, while high resolution unveils finer details. A precedent approach \cite{muller2022instant} introduces a learning-based multi-resolution hash encoding, prioritizing pertinent details autonomously, free from task-specific constraints. In the context of UAV scenes characterized by diverse scales, this methodology can be applied to ensure a comprehensive representation encompassing both substantial structures and minute details. Notably, \textit{prior multi-resolution hash encoding processes features independently at each resolution, neglecting inter-resolution correlations.} To address this, we propose a cross-resolution attention mechanism gauging association degrees between different resolution levels. Following the Transformer model \cite{vaswani2017attention}, Scaled Dot-Product Attention is employed to compute query $Q_{x,y,z,r}$, key $K_{x,y,z,r}$ and value $V_{x,y,z,r}$ on features of distinct resolutions, yielding $E_{x,y,z,r}''$ from the preprocessed point embedding $E_{x,y,z}'$: 
\begin{align}
    % & Q_{x,y,z,r} = W^{(Q)} \cdot E_{x,y,z,r}', \\
    % & K_{x,y,z,r} = W^{(K)} \cdot E_{x,y,z,r}', \\
    % & V_{x,y,z,r} = W^{(V)} \cdot E_{x,y,z,r}', \\
    & \begin{bmatrix} Q_{x,y,z,r} \\ K_{x,y,z,r} \\ V_{x,y,z,r}  \end{bmatrix} = \begin{bmatrix} W^{(Q)} \\ W^{(K)} \\ W^{(V)}  \end{bmatrix} \cdot E_{x,y,z,r}', \\
    & E_{x,y,z,r}'' = \mathrm{Softmax} \left( \frac{ Q_{x,y,z,r} \cdot K_{x,y,z,r}^{\top}}{\sqrt{d_K}} \right) \cdot V_{x,y,z,r},
\end{align}
where $ W^{(Q)}, W^{(K)}, W^{(V)} \in \mathbb{R}^{d_E \times d_K}$ are weight matrices, $d_E$ and $d_K$ is the dimension of the embedding and the attention key. Augmenting the model akin to Transformer \cite{vaswani2017attention}, we introduce a position embedding. This embeds the $e$-th data position at the $r$-th resolution level as follows:
\begin{align}
    & PE_{r,e} = \left\{ 
    \begin{array}{lr}
         \sin \left( \frac{r}{10000^{e/d_{E}}} \right)        & e=2\lambda  \\
         \cos \left( \frac{r}{10000^{(e-1)/d_{E}}} \right)    & e=2\lambda + 1
    \end{array}
    \right. 
    , \lambda \in \mathbb{N}
\end{align}
In pursuit of an exhaustive multi-resolution feature, we integrate an extra global-level feature, denoted as $E_{global}$, within the cross-resolution attention module, depicted in Fig.~\ref{fig:framework}. Analogous to the $[CLS]$ token in the Vision Transformer (ViT) \cite{dosovitskiy2020image}, the $E_{global}$ feature serves to concatenate with features across diverse resolution levels, fostering attention connections throughout all levels.

\textbf{Point location grouping.} The aforementioned feature pool, expounded in \S \ref{sec:framework}, discretely reinstates point space features, causing significant disparities among adjacent points. Previous methods \cite{mildenhall2021nerf,turki2022mega,tancik2022block,xiangli2022bungeenerf} calculate continuous features based on positional information. This involves employing a learnable linear transformation $\varphi$ with a weight $W_{\varphi}$ and bias $B_{\varphi}$ to obtain continuous local features for each point, projecting the point location $(x, y, z) \in \mathbb{R}^3$ (see Fig.~\ref{fig:framework}) into a higher-dimensional space $\mathbb{R}^{3 \times 2 \times V}$ using high-frequency functions.
However, inadequate sampling frequency violates Nyquist-Shannon theorem, causing high-frequency signals to fold into low-frequency signals during continuous signal sampling. Hence, we construct a hybrid point location embedding method by grouping the neighbor points into multivariate Gaussian regions, inputting the expected group location $(\mathbb{E}_r(x), \mathbb{E}_r(y), \mathbb{E}_r(z))$ of the region around the point $(x, y, z)$ as query into the learnable linear transformation $\varphi$:
\begin{equation} \label{eq:point}
    E_{x,y,z}^{location} = W_{\varphi} \cdot \left(\sin (2^0 \pi \mathbb{E}_r(x)), ... , \cos (2^S \pi \mathbb{E}_r(z))\right)   + B_{\varphi},
\end{equation}
where $S$ represents the number of $\sin$ and $\cos$ frequency variations, and $\cdot$ denotes the dot product operation. 

\textbf{Volume feature.} In pursuit of holistic component representation, our methodology delves into the bounding volume hierarchy (BVH) algorithm \cite{wald2007ray}. This geometric query accelerator exploits the insight that if a query object avoids volume intersection, it precludes interaction with objects within. Our approach computes an encompassing feature for each UAV environment segment, stored in a learnable volume feature pool $F^{vol}$. The formalized volume feature embedding is expressed as:
\begin{align}
    E_{x,y,z}^{volume} = \Phi \left( F^{vol}, \delta_{x,y,z} \right), F^{vol} \in \mathbb{R}^{C \times d_E},
\end{align}
where $\delta_{x,y,z}$ represents a sampled point located at the coordinates $(x, y, z)$, while $\Phi$ refers to the volume feature selection algorithm, which involves two key steps: first, determining the volume index of this point $\delta_{x,y,z}$ using the BVH approach, and second, selecting volume features based by this index. $C$ denotes the total count of available volumes.

% Those above methods could improve the \textit{\textbf{global structure}} as well as \textit{\textbf{local granularity}} of the generated video, especially for the low-texture surface.

%-------------------------------------------------------------------------
\subsection{Comprehensive Learning Objective} \label{sec:loss}
Although the method above could solve the almost problem of \textit{\textbf{temporal consistency}}, \textit{\textbf{global structure}}, and \textit{\textbf{local granularity}}, there are still some problem when learning a structure where the view perspective for training supervision is limited. Leveraging the spatio-temporal information of the video, we propose a novel loss function to punish those bad rendering results in apparent details and temporal consistency, involving three parts:

\textbf{Color alignment.} The RGB color reconstruction loss is calculated as by the Charbonnier loss \cite{charbonnier1994two} between the RGB values of the reconstructed images and the ground truth images. We introduce a fixed rectification value into it, denoted as $\gamma$, which achieves slightly more stable optimization than the MSE loss used in mip-NeRF \cite{barron2021mip}. Consequently, this loss term is defined as follows:
\begin{align}
    L_c= \frac{1}{3} \Bigg[ \sqrt{(r_{\phi} - \hat{r}_{\phi})^2 + \gamma} + \sqrt{(g_{\phi} - \hat{g}_{\phi})^2 + \gamma} + \sqrt{(b_{\phi} - \hat{b}_{\phi})^2 + \gamma} \Bigg],
\end{align} \label{eq:color_loss}
where $r_{\phi},g_{\phi},b_{\phi}$ represents the rendered RGB value of the sampled view ray $\phi$, and $\hat{r}_{\phi},\hat{g}_{\phi},\hat{b}_{\phi}$ represents their corresponding ground truth from the original input frames.

\textbf{Disparity alignment.} This loss term serves to quantitatively assess the reciprocal depth disparity between two resolutions, providing robustness against introduced noise to mitigate local distortions. Our methodology employs the Earth Mover's Distance (EMD) \cite{rubner2000earth}, rooted in Optimal Transport theory \cite{cuturi2013sinkhorn}, to quantify the depth distribution difference between resolutions. EMD distinguishes itself from conventional metrics by not only scrutinizing value differences at corresponding points but also incorporating transportation costs into its evaluation, emphasizing distribution shapes. This unique characteristic makes EMD proficient at discerning disparities in distributions with an overarching resemblance but nuanced deviations. In spatial contexts, EMD excels at shape-based comparisons, particularly when analyzing density distributions. In comparing and aligning spatial data point distributions, EMD's virtue lies in its ability to consider both spatial arrangement and structural nuances. The formula expressing the disparity alignment loss function is denoted as::
\begin{align} \label{eq:L_d}
    L_d=\sum_{\delta_i,\delta_j\in \phi} \sqrt{\left(\sum_{0 \leq j \leq i} d_{\delta_j} - \sum_{0 \leq j \leq i} \hat{d}_{\delta_j} \right)^2 + \gamma} ,
\end{align}
where $d_{\delta_j}$ represents the disparity value at the $j$-th point $\delta_j$ on the sampled view ray $\phi$. The cumulative distribution function $\sum_{0 \leq j \leq i} d_{\delta_j}$ is computed for each point $\delta_i$ along this ray, encompassing all points along the sampled view ray $\phi$ to portray the overall disparity distribution of $\phi$.

% \textbf{Total variance loss.}
\textbf{Smoothness consistency.} Outlined in \S \ref{sec:space_estimation}, our methodology involves spatial subdivision based on camera parameters, potentially yielding inconsistencies among neighboring blocks. This loss term assesses the bounded variation of functions relative to adjacent octree nodes, promoting uniform densities and colors at their intersections. Employing the bounded variation function from image restoration \cite{rudin1994total}, suitable for non-continuous distributions, the feature difference between points from adjoining octree nodes is computed as:
\begin{equation}
    L_s = \frac{1}{N} \sum_{(x,y,z)\in \xi} \Big|\Big| F_{x,y,z}^{(\varphi_0)}  - F_{x,y,z}^{(\varphi_1)} \Big|\Big|.
\end{equation}
We randomly sample $N$ edge points on the borders of the octree nodes, this set of points is denoted as $\xi$. For each same edge point with position $(x,y,z)\in \xi$, their features $F_{x,y,z}^{(\varphi_0)}$ and $F_{x,y,z}^{(\varphi_1)}$ are fetched from two different warping functions $\varphi_0$ and $\varphi_1$ conditioned on their corresponding neighboring octree nodes.

\textbf{Final loss.} The final loss is the sum of the above loss terms:
\begin{equation} \label{eq:final_loss}
    L_{total} = \alpha_1 L_c + \alpha_2 L_d + \alpha_3 L_s,
\end{equation}
where $\alpha_1, \alpha_2, \alpha_3$ are the weights for those loss terms. By training the network with the joint supervision of these three loss functions, it not only guarantees comprehensive and detailed rendering quality in the RGB field but, more crucially, maintains the coherence and consistency of the generated 3D model across spatial and temporal dimensions. This aspect has been overlooked by current NeRF methods, with a specific focus on addressing UAV videos that encompass diverse indoor and outdoor multi-scale scenes.

% Those above methods could improve the \textit{\textbf{temporal consistency}} as well as \textit{\textbf{global structure}} of the generated video.

\setlength{\intextsep}{3pt}
\begin{wrapfigure}{r}{0.61\textwidth}
  \centering
   \includegraphics[width=1.0\linewidth]{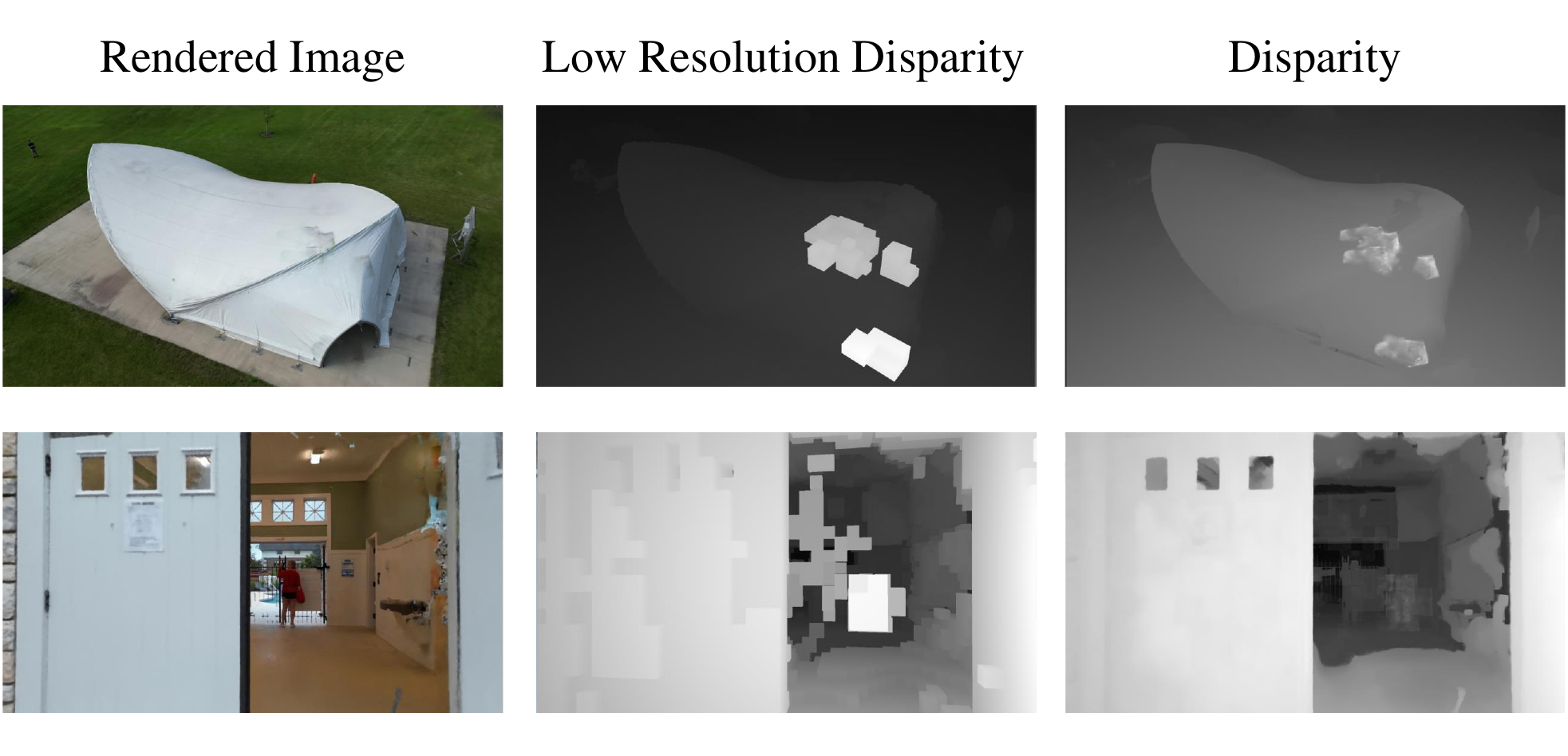}
   \caption{\textbf{Illustration of disparity alignment.} The absence of local granularity results in a substantial difference in reciprocal depth disparity between two resolutions (see \S \ref{sec:loss}).}
   \label{fig:more_disparity_loss}
\end{wrapfigure}

\section{Experiments}

%-------------------------------------------------------------------------
\subsection{UAV Dataset Collection} \label{sec:uav_dataset}
We construct a new benchmark for 3D reconstruction on UAV videos (called the \textit{UAV dataset}). The UAV dataset is captured by DJI mini3 pro, contains ten unbounded scenes, including sky, grassland and a lot of buildings. Each scene is captured by a UAV, which first flies around the building in the sky, and then lands and moves into the building through the door. Each scenario video encompasses a temporal span ranging from 3 to 7 minutes. Simultaneously, the operational altitudes of the UAV manifest a diverse range, extending from 40 to 170 meters. The scenes captured include \textit{market}, \textit{garden}, \textit{park}, \textit{foundation}, \textit{library}, \textit{etc.}, all of which are recorded in New York City, USA.
In the exposition of our aggregated dataset pertaining to UAV, as elucidated in \S \ref{sec:uav_dataset}, an exhaustive calibration procedure unfolds, probing into both internal and external dimensions. This meticulous endeavor is centered upon the precise determination of intrinsic and extrinsic parameters, firmly grounded in a predetermined image size of $3024 \times 4032$ pixels.
The scrutiny of camera intrinsics entails a discerning analysis of critical properties, encompassing focal length, principal point, radial distortion, skew, and the intrinsic matrix. The focal length for the UAV camera, a pivotal intrinsic parameter essential for accurately mapping pixel coordinates to real-world spatial dimensions, is computed to be $(2.858 \times 10^3, 2.845 \times 10^3)$ in our experimental investigations. 
% This meticulously calibrated framework forms the cornerstone for a nuanced comprehension of the intricacies inherent in the UAV camera, thereby fortifying the foundation for subsequent applications in the domains of UAV-based imaging and mapping.

\begin{figure}[t]
  \centering
   \includegraphics[width=0.96\linewidth]{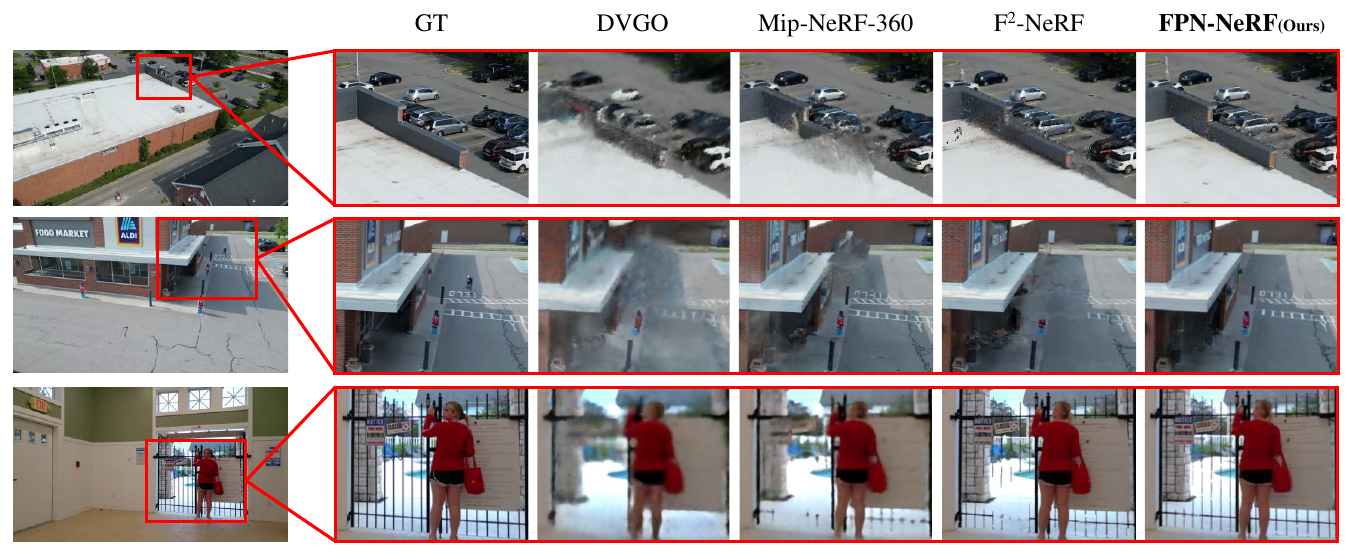}
   \caption{\textbf{Qualification comparison results with several SOTA methods.} It can be seen that the synthesized frames of DVGO is blurred due to their limited resolutions to represent such a long trajectory. The results of Mip-NeRF-360 and F$^2$-NeRF have local noise and distortion due to its unbalanced scene space organization. In comparison, our FPV-NeRF takes advantage of the adaptive space subdivision and considers scene features in different scale to fully exploit the global-local representation capacity.}
   \label{fig:comparison}
   \vspace{-0.5cm}
\end{figure}

%-------------------------------------------------------------------------
\vspace{-0.3cm}
\subsection{Experimental Settings}
\vspace{-0.2cm}
All training experiments are done on a single NVIDIA A6000 GPU. Implementation details of our FPV-NeRF and the comparable baselines are following:

\textbf{Implementation details.}  The point location grouping factor $S$ is set to be 10 in Eq.~\ref{eq:point}. The weights in final loss is fixed to $\alpha_1=1.0, \alpha_2=0.001, \alpha_3=0.1$ in Eq.~\ref{eq:final_loss}. The attention The tiny MLP in scene encoder has one hidden layer of width 64 to get the scene features and the volume densities, while the rendering MLP has two hidden layers of width 64 to get the RGB colors (\S \ref{sec:framework}). For the feature pool training, we follow a similar setting to Instant-NGP \cite{muller2022instant} and use the hash table with 16 levels, each level contains a certain number of feature vectors with dimension of 2. On our UAV dataset, each level contains $2^{21}$ feature vectors and are trained for 80k steps. On the other datasets, each level contains $2^{19}$ and are trained for 20k steps. We train the network with Adam optimizer, whose learning rate linearly grows from zero to $1 \times 10^{-1}$ in the first 1k steps and then decay to $1 \times 10^{-2}$ at the end of training with cosine scheduling, with a batch size of 262144. We adhere to the widely accepted training and testing configurations, randomly selecting 12.5\% of all keyframes from the UAV video for testing images, while the remaining frames constitute the training set. To quantitatively evaluate the novel view synthesizing quality, we utilize three metrics: PSNR, SSIM, and LPIPS$_{(\mathrm{VGG})}$ \cite{zhang2018unreasonable}. As shown in Fig.~\ref{fig:more_disparity_loss}, the low-resolution disparity in \S \ref{sec:loss} adopts reconstruction results from the first octree level, visibly revealing blocks that are positioned inaccurately but exhibit relative transparency.

\textbf{Baselines.} We compare our FPV-NeRF with the state-of-the-art NeRF methods, including (1) the \textit{MLP-based methods}: NeRF++  \cite{zhang2020nerf++}, Mip-NeRF \cite{barron2021mip}, Mip-NeRF 360 \cite{barron2022mip}, and Tri-MipRF \cite{hu2023tri}; (2) the \textit{voxel-based methods}: Plenoxels \cite{fridovich2022plenoxels}, DVGO \cite{sun2022direct}, TensoRF \cite{chen2022tensorf}, Instant-NGP \cite{muller2022instant}, F$^2$-NeRF \cite{wang2023f2}, Mega-NeRF \cite{turki2022mega} and Neo 360 \cite{irshad2023neo}. Note that Instant-NGP \cite{muller2022instant} uses a CUDA implementation while we adopt a LibTorch \cite{paszke2019pytorch} implementation, thus faster than ours.

\begin{table}[t]
  \begin{minipage}[T]{0.49\linewidth}
  \caption{\textbf{Main results on our proposed UAV dataset.}}
  \label{tab:sota_uav_dataset}
  \vspace{-2ex}
  \centering
  \scalebox{0.72}{
  % \begin{tabular}{@{}lcccc@{}}
  \begin{tabular}{l|c|ccc}
    \hline \thickhline
    \rowcolor{mygray}
    Method & Tr. time & PSNR$_\uparrow$ & SSIM$_\uparrow$ & LPIPS$_{\mathrm{(VGG)}\downarrow}$ \\
    \hline \hline
    NeRF++ \cite{zhang2020nerf++} & days & 22.67 & 0.695 & 0.421 \\
    Mip-NeRF 360 \cite{barron2022mip} & days & 25.22 & 0.748 & 0.380 \\
    Zip-NeRF \cite{barron2023zip} & days & 24.78 & 0.703 & 0.413 \\
    \midrule
    Plenoxels \cite{fridovich2022plenoxels} & 8.4h & 19.06 & 0.624 & 0.475 \\
    DVGO \cite{sun2022direct} & 7.3h & 21.78 & 0.683 & 0.419 \\
    Instant-NGP \cite{muller2022instant} & 3.9h & 23.17 & 0.710 & 0.392 \\
    F$^2$-NeRF \cite{wang2023f2} & 6.1h & 24.61 & 0.735 & 0.386 \\
    Mega-NeRF \cite{turki2022mega} & 9.8h & 23.95 & 0.726 & 0.391 \\
    \textbf{FPV-NeRF$_{\mathrm{(Ours)}}$} & 5.5h & \textbf{26.83} & \textbf{0.769} & \textbf{0.364} \\
    \bottomrule
  \end{tabular}
  }
  \vspace{3ex}
  \caption{\textbf{Results on previous NeRF-360-V2 \cite{barron2022mip} dataset.}}
  \label{tab:sota_360_dataset}
  \vspace{-2ex}
  \centering
  \scalebox{0.71}{
  \begin{tabular}{l|c|ccc}
    \hline \thickhline
    \rowcolor{mygray}
    Method & Tr. time & PSNR$_\uparrow$ & SSIM$_\uparrow$ & LPIPS$_{\mathrm{(VGG)}\downarrow}$ \\
    \hline \hline
    NeRF++ \cite{zhang2020nerf++} & hours & 26.21 & 0.729 & 0.348 \\
    Mip-NeRF 360 \cite{barron2022mip} & hours & 28.94 & 0.837 & 0.208 \\
    Zip-NeRF \cite{barron2023zip} & hours & 28.25 & 0.822  & 0.198 \\
    \midrule
    Plenoxels \cite{fridovich2022plenoxels} & 22m & 23.35 & 0.651 & 0.471 \\
    DVGO \cite{sun2022direct} & 16m & 25.42 & 0.695 & 0.429 \\
    Instant-NGP \cite{muller2022instant} & 6m & 26.24 & 0.716 & 0.404 \\
    F$^2$-NeRF \cite{wang2023f2} & 14m & 26.39 & 0.746 & 0.361 \\
    Neo 360 \cite{irshad2023neo} & 19m & 25.88 & 0.707 & 0.412 \\
    \textbf{FPV-NeRF$_{\mathrm{(Ours)}}$} & 12m & \textbf{26.42} & \textbf{0.758} & \textbf{0.359} \\
    \bottomrule
  \end{tabular}
  }
  \vspace{3ex}
  \caption{\textbf{Results on previous LLFF \cite{mildenhall2019local} dataset.}}
  \label{tab:sota_llff_dataset}
  \vspace{-2ex}
  \centering
  \scalebox{0.71}{
  \begin{tabular}{l|c|ccc}
    \hline \thickhline
    \rowcolor{mygray}
    Method & Tr. time & PSNR$_\uparrow$ & SSIM$_\uparrow$ & LPIPS$_{\mathrm{(VGG)}\downarrow}$ \\
    \hline \hline
    NeRF++  \cite{zhang2020nerf++} & hours & 26.50 &0.811 & 0.250 \\
    Mip-NeRF  \cite{barron2021mip} & hours & 26.60 & 0.814 & 0.246 \\
    \midrule
    Plenoxels \cite{fridovich2022plenoxels} & 17m & 26.29 & 0.822 & 0.210 \\
    DVGO \cite{sun2022direct} & 11m & 26.34 & 0.838 & 0.197 \\
    TensoRF \cite{chen2022tensorf}  & 48m & 26.73 & 0.839 & 0.204 \\
    Instant-NGP \cite{muller2022instant} & 6m & 25.09 & 0.758 & 0.267 \\
    F$^2$-NeRF \cite{wang2023f2} & 13m & 26.54 & 0.844 & 0.189 \\
    Tri-MipRF \cite{hu2023tri} & 20m & 25.61 & 0.792 & 0.199 \\
    \textbf{FPV-NeRF$_{\mathrm{(Ours)}}$} & 10m & \textbf{26.74} & \textbf{0.852} & \textbf{0.185} \\
    \bottomrule
  \end{tabular}
  }
  \end{minipage}
  \hspace{0.01\linewidth}
  \begin{minipage}[T]{0.49\linewidth}
  \caption{\textbf{Comparison of different camera space estimation methods (\S \ref{sec:space_estimation}) on UAV dataset.} We first assess the time complexity $O$ of various trajectory prediction (TP) methods: sequential matching (Seq.), which matches nearby frames, and exhaustive matching (Exh.), which matches all frames in the video. Additionally, we compare the performance between our method and the scenario without block subdivision (BS).}
  \label{tab:ablation_space_estimation}
  \renewcommand{\arraystretch}{1.23}
  \centering
  \scalebox{0.75}{
  % \begin{tabular}{@{}lcccc@{}}
  \begin{tabular}{l|c|ccc}
    \hline \thickhline
    \rowcolor{mygray}
    Method & $O$ of TP & PSNR$_\uparrow$ & SSIM$_\uparrow$ & LPIPS$_{\mathrm{(VGG)}\downarrow}$ \\
    \hline \hline
    Seq.  & $O(T^2 \cdot n)$ & \textbf{26.83} & \textbf{0.769} & \textbf{0.364} \\
    \midrule
    Exh.  & $O(n^2)$ & 25.31 & 0.718 & 0.382 \\
    Seq. (w/o BS) & $O(T^2 \cdot n)$ & 21.60 & 0.605 & 0.446 \\
    
    \bottomrule
  \end{tabular}
  }
  \vspace{7.5ex}
  \caption{\textbf{Effect of global-local encoder (\S \ref{sec:encoder}) on UAV dataset.} We systematically exclude the cross-resolution attention, grouped point location embedding $E_{x,y,z}^{location}$, and volume embedding $E_{x,y,z}^{volume}$ from our proposed global local scene encoder, individually assessing their impact.}
  \label{tab:ablation_encoder}
  \centering
  \scalebox{0.78}{
  % \begin{tabular}{@{}lcccc@{}}
  \begin{tabular}{l|c|ccc}
    \hline \thickhline
    \rowcolor{mygray}
    Method & Tr. time & PSNR$_\uparrow$ & SSIM$_\uparrow$ & LPIPS$_{\mathrm{(VGG)}\downarrow}$ \\
    \hline \hline
    -  & 5.5h & \textbf{26.83} & \textbf{0.769} & \textbf{0.364} \\
    \midrule
    w/o Attention  & 4.2h & 22.47 & 0.691 & 0.420 \\
    w/o $E_{x,y,z}^{location}$  & 5.0h & 24.26 & 0.713 & 0.405 \\
    w/o $E_{x,y,z}^{volume}$ & 5.3h & 25.71 & 0.742 & 0.388 \\ 
    
    \bottomrule
  \end{tabular}
  }
  \end{minipage}
  \vspace{-0.3cm}
\end{table}

\begin{figure}
\begin{minipage}[t]{0.42\linewidth}
  \centering
   \includegraphics[width=1.0\linewidth]{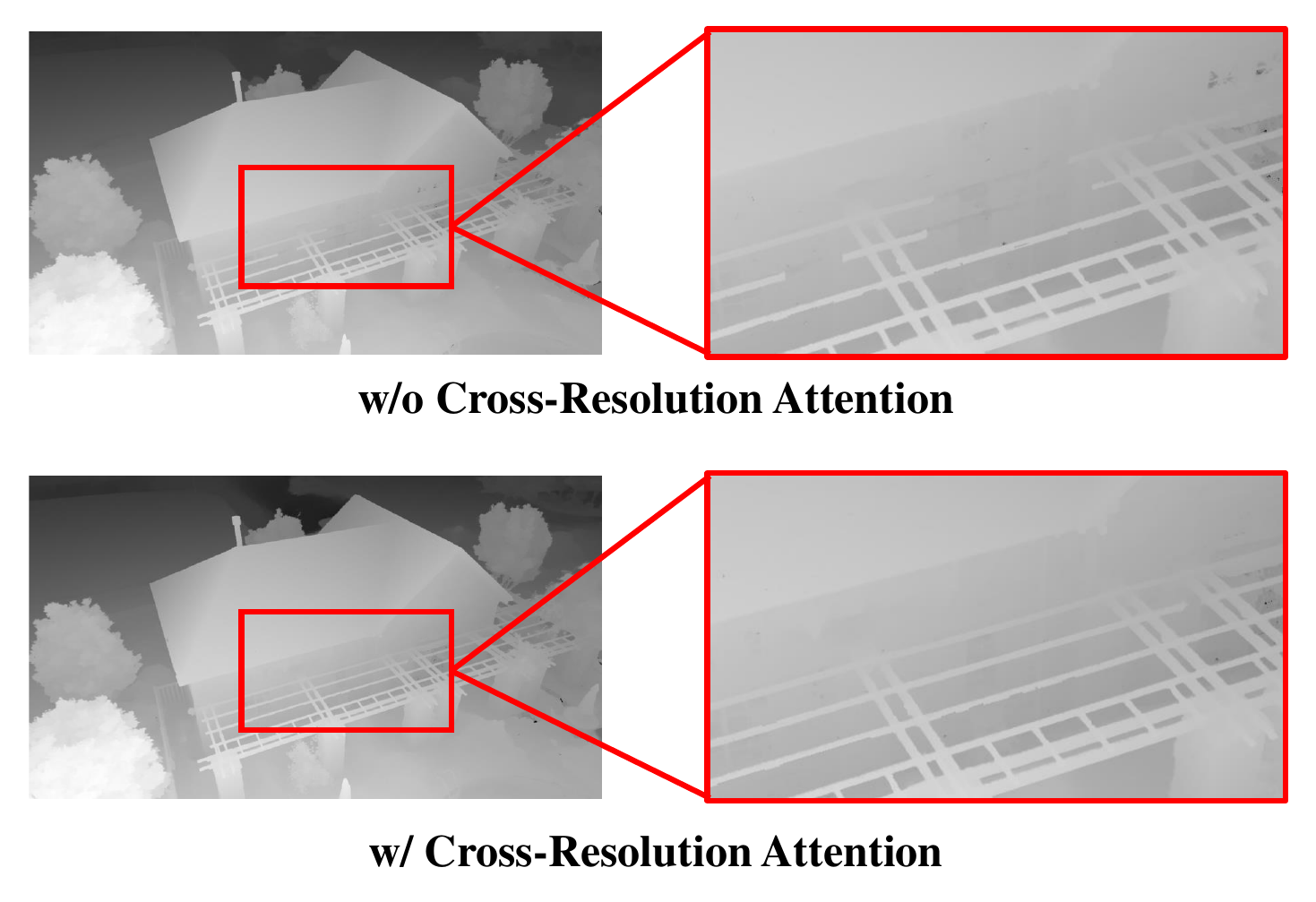}
   \caption{\textbf{Ablation study results on Cross-Resolution Attention,} which plays an important role in global-local encoder (see \S \ref{sec:encoder}).}
   \label{fig:disp}
\end{minipage}
\hspace{0.01\linewidth}
\begin{minipage}[t]{0.55\linewidth}
  \centering
   \includegraphics[width=1.0\linewidth]{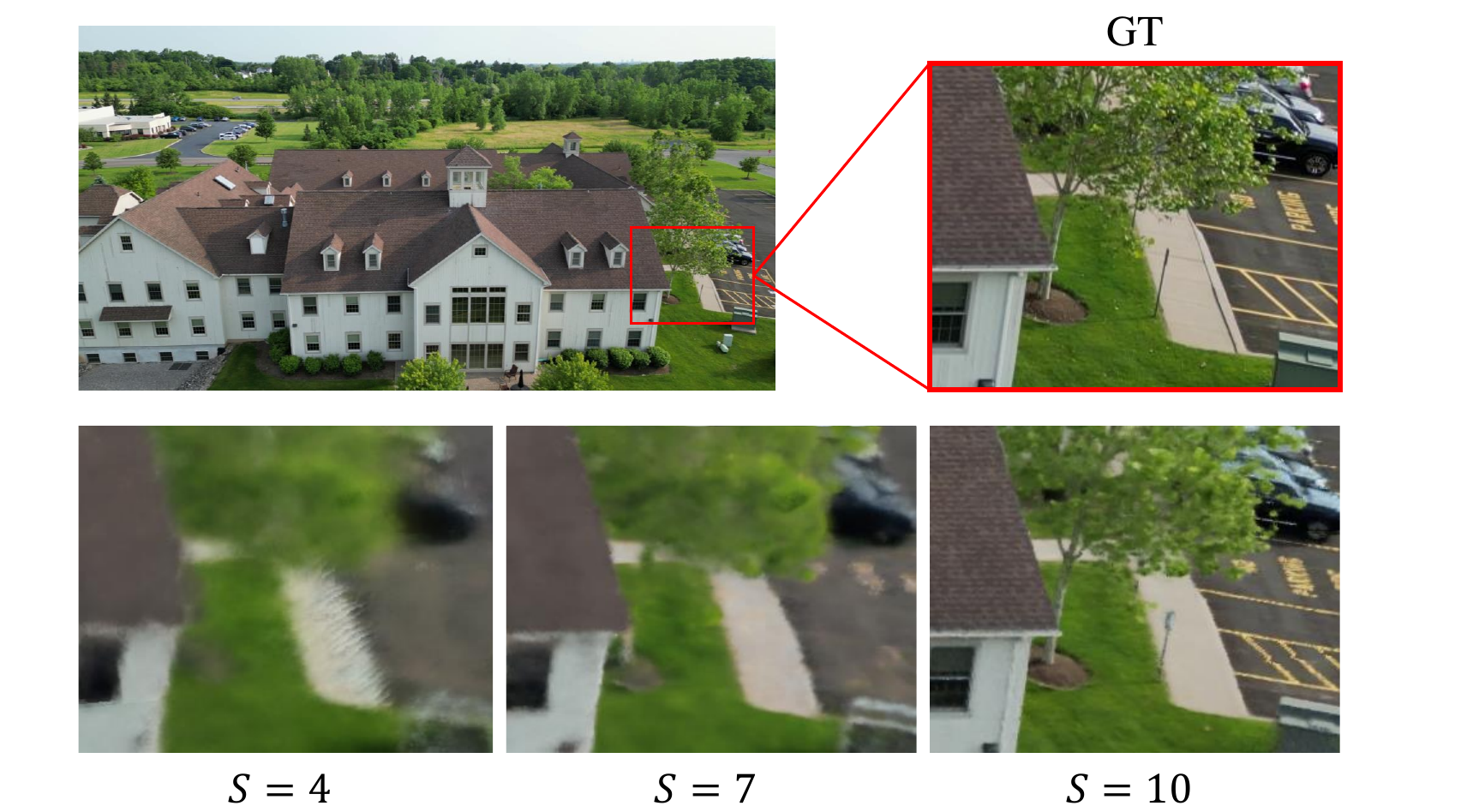}
   \caption{\textbf{Comparison of different point location grouping parameter $S$ in Eq.~\ref{eq:point}.} Larger $S$ exhibits more low-frequency and high-frequency components of the scene (see \S \ref{sec:encoder}). }
   \label{fig:point_location}
\end{minipage}
\vspace{-0.5cm}
\end{figure}

\vspace{-0.3cm}
\subsection{Comparison with SOTA} \label{sec:comparison_sota}
\vspace{-0.2cm}
We first report the quantitative comparisons on our proposed UAV dataset and then further validate our method on previous popular NeRF datasets. 

\textbf{Results on the proposed UAV dataset.} As shown in Table~\ref{tab:sota_uav_dataset} and Fig.~\ref{fig:comparison}, the synthesized frames of DVGO \cite{sun2022direct} is blurred due to their limited resolutions to represent such a long trajectory, the results of Mip-NeRF 360 \cite{barron2022mip} and F$^2$-NeRF \cite{wang2023f2} look shaper but have local noise and distortion due to its unbalanced scene space organization. In comparison, our FPV-NeRF takes advantage of the adaptive space subdivision and considers scene features in different scale to fully exploit the global-local representation capacity. Specifically, FPV-NeRF surpasses the prior leading method F$^2$-NeRF \cite{wang2023f2} by a margin of 2.22 PSNR, 0.034 SSIM, and 0.022 LPIPS$_{\mathrm{(VGG)}\downarrow}$. Additionally, it outperforms the former second-best method, Instant-NGP \cite{muller2022instant}, by a margin of 3.66 PSNR, 0.059 SSIM, and 0.028 LPIPS$_{\mathrm{(VGG)}\downarrow}$. For each view in our UAV dataset the average inference times of our FPV-NeRF and Mega-NeRF \cite{turki2022mega} are 13.7s and 21.6s respectively.

\textbf{Results on previous datasets.} To show the compatibility of FPV-NeRF, we also evaluate our method on two widely-used datasets with two kinds of specialized camera trajectories in Fig.~\ref{fig:intro}(a) and Fig.~\ref{fig:intro}(b): (1) NeRF-360-V2 dataset \cite{barron2022mip}, which contains seven unbounded 360-degree inward-facing scenes. (2) LLFF dataset \cite{mildenhall2019local}, which contains eight real unbounded forward-facing scenes recorded by mobile phone. As illustrated in Table~\ref{tab:sota_360_dataset} and Table~\ref{tab:sota_llff_dataset}, FPV-NeRF consistently attains comparable outcomes to leading voxel-based approaches on both datasets. Specifically, it exhibits improvements of 0.03-3.07 PSNR and 0.012-0.107 SSIM on the NeRF-360-V2 dataset, 0.01-1.65 PSNR and 0.008-0.094 SSIM on the LLFF dataset.

\begin{wraptable}{t}{0.6\linewidth}
  \vspace{-15pt}
  \caption{\textbf{Impact of different weights for loss terms (\S \ref{sec:loss}) on UAV dataset.} We fix the weight $\alpha_1=1.0$ for $L_c$  and adjust weights for $L_d$ and $L_s$ in Eq.~\ref{eq:final_loss}. All results are reported on PSNR.}
  \label{tab:ablation_loss}
  \vspace{10pt}
  \centering
  \setlength{\tabcolsep}{3mm}{
  \scalebox{0.71}{
  % \begin{tabular}{@{}lcccc@{}}
  \begin{tabular}{l|cc>{\columncolor{mygray}}ccc}
    \hline \thickhline
    \rowcolor{mygray}
    \diagbox{$\alpha_3$}{$\alpha_2$} & 0 & $10^{-4}$ & $10^{-3}$ & $10^{-2}$ & $10^{-1}$ \\
    \hline \hline
    0  & 21.98 & 22.06 & 23.41 & 22.19 & 20.52 \\
    $10^{-2}$  & 22.37 & 23.85 & 25.20 & 23.13 & 21.67 \\
    \rowcolor{mygray}
    $10^{-1}$ & 23.52 & 25.34 & \textbf{26.83} & 24.78 & 22.05 \\ 
    $10^{0}$ & 21.01 & 22.29 & 23.36 & 21.67 & 19.56 \\ 
    
    \bottomrule
  \end{tabular}
  }}
\end{wraptable}

%-------------------------------------------------------------------------
\subsection{Ablation Study} \label{sec:ablation_study}
We conduct ablation studies on our proposed UAV dataset to prove the effect of each module we propose.

\textbf{Different camera space estimation method.} Employing exhaustive matching results (see Table~\ref{tab:ablation_space_estimation}) in a quadratic time complexity increment of $O(n^2)$ for trajectory prediction (\S \ref{sec:space_estimation}). Conversely, performance exhibits a decline of 1.52 PSNR, 0.051 SSIM, and 0.018 LPOS${_\mathrm{VGG}}$. Furthermore, our approach surpasses the non-subdivided block counterpart by 5.23 PSNR and 0.164 SSIM.
% , and 0.082 LPOS${_\mathrm{VGG}}$.

\begin{wrapfigure}{r}{0.65\textwidth}
  \centering
  \vspace{-5pt}
   \includegraphics[width=1.0\linewidth]{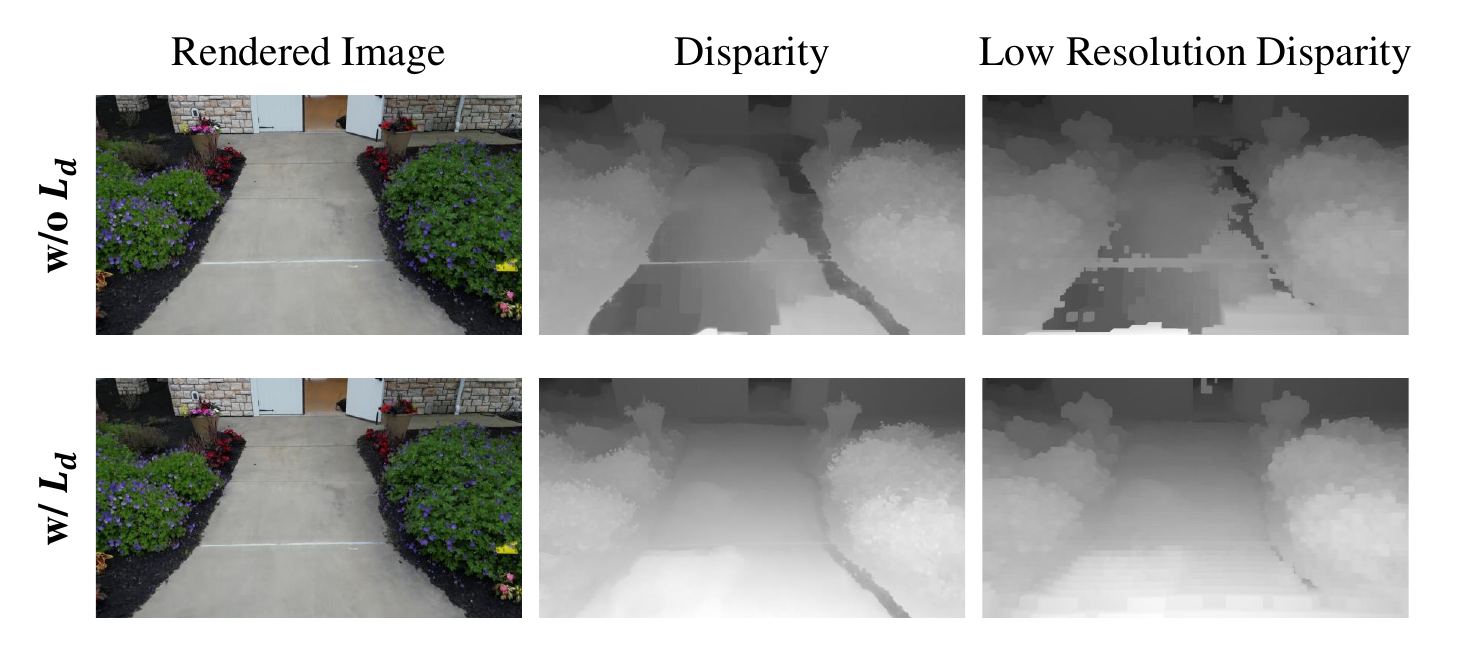}
   \caption{\textbf{Ablation study results on $L_d$ in Eq.~\ref{eq:L_d}.} The RGB image that has been processed appears normal at first glance, but a closer examination reveals that the disparity (relative to the depth) indicates that the road surface is not smooth, with deep caves. Disparity from different resolution levels can reflect this problem, whereas the color loss cannot reveal it (see \S \ref{sec:loss}).}
   \label{fig:disp_loss}
\end{wrapfigure}

\textbf{Effect of global-local scene encoder.} As depicted in Table~\ref{tab:ablation_encoder}, the exclusion of cross-resolution attention, point location grouping, and volume embedding from our global-local encoder (\S \ref{sec:encoder}) leads to a reduction in PSNR performance by 1.12-4.36 and a decrease in SSIM by 0.027-0.078. Additionally, Fig.\ref{fig:disp} illustrates that the inclusion of cross-resolution attention preserves the integrity of complex structures, while Fig.\ref{fig:point_location} demonstrates that employing a larger $S$ in Eq.~\ref{eq:point} enhances the vividness of reconstruction details.

\textbf{Ablation study on learning objective.} As depicted in Table~\ref{tab:ablation_loss}, we systematically investigate varying values of $\alpha_2$ and $\alpha_3$ to gauge the impact of $L_d$ and $L_s$. Our goal is to pinpoint the optimal loss parameters in Eq.~\ref{eq:final_loss} for FPV-NeRF, as detailed in \S \ref{sec:loss}. Setting $\alpha_2=0$ or $\alpha_3=0$ omits $L_d$ or $L_s$ from our loss, causing a PSNR reduction of 4.85. The experimentation reveals that introducing $L_d$ and $L_s$ as supervisory signals at an appropriate order of magnitude enhances FPV-NeRF results. However, caution is warranted, as excessively high weights for $\alpha_2$ and $\alpha_3$ can lead to interference with $L_c$ during the early stages of training. Additionally, we present intuitive comparison results for $L_d$ in Fig.~\ref{fig:disp_loss}. The findings indicate that employing $L_d$ ensures the local granularity of the 3D reconstruction. In contrast, $L_c$ occasionally struggles to accurately restrict the distance between pixels and the camera.

In summary, the temporal consistency is ensured by sequential trajectory method (Table~\ref{tab:ablation_space_estimation}) and smoothness consistency (Table~\ref{tab:ablation_loss}). The local granularity is achieved through disparity alignment and global-local encoding (Fig.\ref{fig:disp}, \ref{fig:point_location} and \ref{fig:disp_loss}). The global structure is ensured through cross-resolution attention, grouped point location and volume embeddings (Table~\ref{tab:ablation_encoder}).

\vspace{-8pt}
\section{Conclusion}
\vspace{-6pt}
%In this work, FPV-NeRF revolutionizes drone spatial perception with a novel NeRF-based approach. It conquers UAV video reconstruction challenges through multi-scale camera estimation, global-local scene encoding, and a robust learning objective. Rigorous experiments showcase FPV-NeRF's superiority, outperforming benchmarks in our carefully curated UAV dataset across diverse trajectories.
In this work, we present FPV-NeRF, a pioneering approach for synthesizing views from limited FPV footage to enhance spatial perception in drone applications. FPV-NeRF addresses UAV video reconstruction challenges with multi-scale camera estimation, a global-local scene encoder, and a comprehensive learning objective focusing on temporal consistency, global structure, and local granularity. Experiments on various trajectories show FPV-NeRF's superiority, surpassing current benchmarks on meticulously curated UAV dataset. Our approach enables reconstructed environments for offline UAV navigation model training and supports UAV tasks like object detection and autonomous navigation.

\bibliographystyle{splncs04}
\bibliography{main}

\end{document}